\documentclass[12pt,3p,times]{elsarticle}

\usepackage{amsmath,amsfonts,amssymb}
\usepackage{amsthm}
\usepackage{mathtools}
\usepackage{bm}
\usepackage{graphicx, epstopdf}
\usepackage[position=top,labelfont=normalfont,textfont=normalfont,singlelinecheck=off,justification=raggedright]{subfig}
\usepackage{subcaption}
\usepackage{floatrow}
\usepackage[dvipsnames]{xcolor}
\usepackage{setspace}
\usepackage{enumerate, etaremune}
\usepackage{booktabs}
\usepackage{tabularx, multirow}
\usepackage{makecell}
\usepackage{diagbox}
\usepackage{array}
\usepackage{framed}
\usepackage{hhline}
\usepackage{url}
\usepackage[unicode,bookmarks=false]{hyperref}
\hypersetup{colorlinks}
\usepackage[toc,page]{appendix}
\usepackage[T1]{fontenc}
\usepackage{cancel}
\usepackage{kotex} 
\usepackage{bookmark}

\usepackage{lineno}

\usepackage[textsize=footnotesize,linecolor=red,backgroundcolor=red!25,bordercolor=red]
  {todonotes}

\usepackage{algorithm, algorithmicx,algpseudocode}
\usepackage{cases}

\newcommand{\eg}{{\it e.g.}}

\newcommand{\etal}{{\it et al.}}

\newcommand{\el}{\mathrm{e}}

\newcommand{\rn}[1]{\uppercase\expandafter{\romannumeral #1\relax}}

\DeclareMathOperator{\grad}{\nabla}
\DeclareMathOperator{\diver}{\nabla\cdot}

\newsavebox{\dotbox}

\theoremstyle{remark}

\newcommand\eatpunct[1]{}

\renewcommand{\arraystretch}{1.3}
\newcolumntype{L}[1]{>{\raggedright\let\newline\\arraybackslash\hspace{0pt}}m{#1}}
\newcolumntype{C}[1]{>{\centering\let\newline\\arraybackslash\hspace{0pt}}m{#1}}
\newcolumntype{R}[1]{>{\raggedleft\let\newline\\arraybackslash\hspace{0pt}}m{#1}}

\allowdisplaybreaks

\biboptions{sort&compress,square,comma,numbers}
\AtBeginDocument{\hypersetup{citecolor=MidnightBlue,linkcolor=MidnightBlue,urlcolor=MidnightBlue}}

\usepackage{etoolbox}
\makeatletter
\patchcmd{\ps@pprintTitle}
{Preprint submitted to}
{~}
{}{}
\makeatother

\begin{document}

\begin{frontmatter}

\title{Deep operator network for surrogate modeling of poroelasticity with random permeability fields}

\author[KAIST]{Sangjoon Park}
\author[NCSU]{Yeonjong Shin}
\author[SNU-CEE,SNU-ICE]{Jinhyun Choo\corref{corr}}
\ead{jinhyun.choo@snu.ac.kr}

\cortext[corr]{Corresponding Author}

\address[KAIST]{Department of Civil and Environmental Engineering, KAIST, Daejeon, South Korea}
\address[NCSU]{Department of Mathematics, North Carolina State University, Raleigh, United States}
\address[SNU-CEE]{Department of Civil and Environmental Engineering, Seoul National University, Seoul, South Korea}
\address[SNU-ICE]{Institute of Construction and Environmental Engineering, Seoul National University, Seoul, South Korea}

\journal{~}


\begin{abstract}
Poroelasticity---coupled fluid flow and elastic deformation in porous media---often involves spatially variable permeability, especially in subsurface systems. 
In such cases, simulations with random permeability fields are widely used for probabilistic analysis, uncertainty quantification, and inverse problems. 
These simulations require repeated forward solves that are often prohibitively expensive, motivating the development of efficient surrogate models. 
However, efficient surrogate modeling techniques for poroelasticity with random permeability fields remain scarce.
In this study, we propose a surrogate modeling framework based on the deep operator network (DeepONet), a neural architecture designed to learn mappings between infinite-dimensional function spaces. 
The proposed surrogate model approximates the solution operator that maps random permeability fields to transient poroelastic responses. 
To enhance predictive accuracy and stability, we integrate three strategies: nondimensionalization of the governing equations, input dimensionality reduction via Karhunen--Lo\'eve expansion, and a two-step training procedure that decouples the optimization of branch and trunk networks.
The methodology is evaluated on two benchmark problems in poroelasticity: soil consolidation and ground subsidence induced by groundwater extraction. 
In both cases, the DeepONet achieves substantial speedup in inference while maintaining high predictive accuracy across a wide range of permeability statistics. 
These results highlight the potential of the proposed approach as a scalable and efficient surrogate modeling technique for poroelastic systems with random permeability fields.
\end{abstract}


\end{frontmatter}


\section{Introduction} 
Poroelasticity describes the interaction between fluid flow and elastic deformation in porous media. 
It provides the modeling basis for many subsurface applications in which groundwater flow and ground deformation are tightly coupled (\eg~\cite{ferronato2008numerical, castelletto2015coupled, chiaramonte2015probabilistic, chang2016injection, settgast2017fully, choo2018large, fan2019basement, zhao2020stabilized, fei2023phase, fei2025crack}).

In subsurface systems, permeability often exhibits strong spatial heterogeneity, spanning several orders of magnitude~\cite{smith1979stochastic, rehfeldt1992field, griffiths1993seepage, elkateb2003overview, baecher2005reliability, fenton2008risk, kang2017improved}. 
This variability plays a critical role in both flow and deformation processes, and therefore often needs to be considered.
A common approach for this purpose is to represent permeability as a spatially correlated random field~\cite{vanmarcke1983random} and propagate its influence through the governing poroelastic equations.
This strategy has been extensively applied in uncertainty quantification for consolidation~\cite{huang2010probabilistic, cheng2017consolidation, wang2022probabilistic}, stochastic modeling of subsidence~\cite{frias2004stochastic, ferronato2006stochastic, wang2015technique, deng2022probabilistic}, and Bayesian inference of settlement and aquifer properties~\cite{alghamdi2021bayesian, tian2022efficient, tian2023data}.

However, random field modeling involves a significant computational challenge that arises from the need to solve the forward poroelastic equations repeatedly over large ensembles of permeability realizations. 
Full-order solvers such as the finite element method (FEM) offer high accuracy and flexibility, but their cost becomes prohibitive in many-query settings, including probabilistic analysis, uncertainty quantification, and inverse problems.

To alleviate this cost, surrogate models are often employed to replace full-order solvers with computationally efficient approximations, and they have demonstrated success in a wide range of mechanics applications~\cite{xiu2002modeling, bui2008model, sudret2014polynomial, peherstorfer2015dynamic, swischuk2019projection, sharma2024physics}. 
In poroelasticity and related problems, polynomial chaos expansion (PCE) surrogates have primarily been used to assess the influence of spatial variability in the coefficient of consolidation on global measures such as the average degree of consolidation~\cite{bong2014probabilistic, bong2018efficient}. 
Projection-based reduced-order models (ROMs), particularly those based on proper orthogonal decomposition, have also been applied to accelerate reservoir simulations~\cite{jin2020reduced}.

Nevertheless, when the input field, such as permeability, is high-dimensional and spatially random, and the outputs of interest are full spatiotemporal fields, even surrogate approaches can become computationally demanding~\cite{mo2019deep, kadeethum2021framework, bahmani2025neural}. 
This limitation highlights the need for surrogate models that scale efficiently with both input and output dimensionality.

Recent advances in deep learning have introduced new surrogate modeling strategies for poroelasticity, capable of capturing nonlinear relationships between inputs and outputs in high-dimensional settings. 
Convolutional neural networks, particularly U-Net architectures, have been applied to predict displacement and pressure fields under spatially random permeability~\cite{tang2022deepA, tang2022deepB, han2024surrogate, han2025accelerated}. 
Continuous conditional generative adversarial networks (CcGANs) have also been employed to model poroelastic responses in heterogeneous porous media~\cite{kadeethum2022continuous}. 

Despite this progress, current deep learning surrogates for poroelasticity still face key limitations. 
Grid-based inputs in U-Net architectures inflate the dimensionality of already complex permeability fields, increasing training costs and hindering generalization. 
GAN-based surrogates, although expressive, are often unstable and difficult to train robustly, particularly when targeting full spatiotemporal predictions~\cite{saxena2021generative}.

In this work, we develop a surrogate modeling framework for poroelasticity with spatially variable permeability, based on the deep operator network (DeepONet)~\cite{lu2021learning}. 
DeepONet is a neural architecture designed to approximate nonlinear operators between function spaces, supported by a universal approximation theorem for operators~\cite{chen1995universal}. 
Within this framework, we use DeepONet to learn the solution operator that maps permeability field realizations to the corresponding transient displacement and pressure fields.

To enhance generalization and training stability, we incorporate three key strategies. 
First, the governing equations and physical variables are nondimensionalized to reduce scale disparities. 
Second, the input dimensionality is reduced through a truncated Karhunen--Lo\'eve (K--L) expansion of the log-permeability field. 
Third, a two-step training procedure is adopted to decouple the optimization of the branch and trunk networks, following recent advances in DeepONet training~\cite{lee2024training}.

The surrogate model is evaluated on two benchmark problems in poroelasticity: soil consolidation  and ground subsidence induced by groundwater extraction.
For each case, training datasets are generated from finite element simulations. 
In both problems, the surrogate achieves high predictive accuracy and delivers orders-of-magnitude speedup over conventional finite element methods.

\section{Poroelasticity with random permeability fields} 
\label{sec:poroelasticity}

In this work, we consider the $\bm{u}$--$p$ formulation for poroelasticity, where coupled elastic deformation and fluid flow are described by two primary variables: the solid displacement vector $\bm{u}$ and the pore pressure $p$. 
The $\bm{u}$--$p$ formulation is the standard approach for poroelasticity and its extensions (\eg~\cite{borja1995mathematical, white2008stabilized, sun2013stabilized, borja2016cam, choo2016hydromechanical, sun2017mixed, choo2019stabilized}). 
In the following, we adopt the excess pore pressure form of this formulation, where $p$ denotes the excess pore pressure---the deviation from a steady-state hydrostatic distribution.

Consider a porous medium saturated by a single incompressible fluid.
The medium occupies the spatial domain $\Omega \subset \mathbb{R}^{d}$, where $d$ denotes the number of spatial dimensions. 
The boundary of the domain, $\partial \Omega$, is partitioned into displacement (Dirichlet) boundaries $\partial \Omega_{\bm{u}}$ and traction (Neumann) boundaries $\partial \Omega_{\bm{t}}$ for the solid deformation field, and into pore pressure boundaries $\partial \Omega_{p}$ and flux boundaries $\partial \Omega_{q}$ for the fluid flow field.
These boundary subsets are assumed to satisfy the standard non-overlapping and closure conditions.
The temporal domain is defined as $\mathcal{T} \coloneqq (0, T]$ with $T > 0$, and the full spatiotemporal domain is given by $\Omega \times \mathcal{T}$.

Without loss of generality, we assume that the porous medium is quasi-static, undergoes infinitesimal deformation, contains no mass sources or sinks, and is composed of incompressible solid grains. 
We also postulate that the effective stress principle holds.
Under these assumptions, the governing equations of poroelasticity are expressed as
\begin{align}
    \diver\bm{\sigma}' - \grad p + \rho_{b} \bm{g} = \bm{0} &\quad\text{(linear momentum balance)},\\
    \diver \bm{\dot{u}} + \diver \bm{q} = 0 &\quad\text{(mass balance)}, 
    \label{eq:balance-eqs} 
\end{align}
where 
$\bm{\sigma}'$ is the effective stress,
$\rho_{b}$ is the buoyant density of the medium, 
$\bm{g}$ is the gravitational acceleration,
and $\bm{q}$ is the superficial fluid velocity.
Note that the buoyant density appears in the momentum balance equation because the formulation is expressed in terms of excess pore pressure.

To close the formulation, we introduce constitutive relations for the solid and fluid responses. 
Assuming the solid deformation is linear elastic, the effective stress--strain relation is expressed as
\begin{equation}
    \bm{\sigma}' = \mathbb{C}^{\el}:\bm{\epsilon}, 
    \label{eq:linear-elasticity} 
\end{equation} 
where $\bm{\epsilon}$ is the infinitesimal strain, defined as
\begin{equation}
    \bm{\epsilon} := \dfrac{1}{2}\left(\grad\bm{u} + \grad^{\mathsf T}\bm{u}\right),
    \label{eq:strain} 
\end{equation} 
and $\mathbb{C}^{\el}$ is the fourth-order elasticity (stiffness) tensor.
For an isotropic elastic material, $\mathbb{C}^{\el}$ is characterized by two independent elasticity parameters, such as Young’s modulus $E$ and Poisson’s ratio $\nu$.
Assuming Darcy's law governs fluid flow, the superficial fluid velocity is given by
\begin{equation}
    \bm{q} = -\frac{k}{\mu_f} \grad p,
\label{eq:darcy-law}
\end{equation}
where
$k$ is the intrinsic permeability,
$\mu_f$ is the dynamic viscosity of the pore fluid, and
$\rho_f$ is the fluid density.
Note that the gravitational contribution is omitted from Eq. \eqref{eq:darcy-law} because $p$ denotes excess pore pressure.

To model spatial variability in permeability, we represent the log-permeability $\kappa := \ln(k)$ as a Gaussian random field. 
This choice is widely used in subsurface modeling due to the high variability and positivity of permeability (\eg~\cite{huang2010probabilistic, cheng2017consolidation, bong2018efficient, deng2022probabilistic}). 
Assuming second-order stationarity, $\kappa$ is fully characterized by a mean $\mu_\kappa$ and a covariance function $C_\kappa(\bm{x}_1, \bm{x}_2)$, typically parameterized by a standard deviation $\sigma_\kappa$ and a correlation length $l$.
We assume that the covariance type is prescribed, and that the statistical parameters are obtained from measurement data (\eg, in-situ soil tests \cite{degroot1993estimating, firouzianbandpey2014spatial, liu2017integrated, liu2018characterising}).
Accordingly, the spatially varying permeability is expressed as
\begin{equation} 
    k(\bm{x}; \omega) = \exp\left(\mu_\kappa + g_\kappa(\bm{x}; \omega)\right),
    \label{eq:permeability field}
\end{equation}
where $\omega \in \Omega$ is an element of the sample space $\Omega$, and $g_\kappa$ is a zero-mean Gaussian field with covariance function $C_\kappa$.

The boundary conditions are specified as
\begin{align}
    \bm{u}(\bm{x}, t) = \hat{\bm{u}}(\bm{x}, t) &\quad\text{on}\;\; \partial \Omega_{\bm{u}} \times \mathcal{T}, \\
    \bm{\sigma}(\bm{x}, t) \bm{n}(\bm{x}) = \hat{\bm{t}}(\bm{x}, t) &\quad\text{on}\;\; \partial \Omega_{\bm{t}} \times \mathcal{T}, \\
    p(\bm{x}, t) = \hat{p}(\bm{x}, t) &\quad\text{on}\;\; \partial \Omega_{p} \times \mathcal{T}, \\
    \bm{q}(\bm{x}, t) \cdot \bm{n}(\bm{x}) = \hat{q}(\bm{x}, t) &\quad\text{on}\;\; \partial \Omega_q \times \mathcal{T},
    \label{eq:boundary-condition}
\end{align}
where
$\bm{x} \in \Omega$ is the position vector,
$t \in \mathcal{T}$ is the time variable, and
$\bm{n}$ is the outward unit normal on the boundary.
The quantities $\hat{\bm{u}}$, $\hat{\bm{t}}$, $\hat{p}$, and $\hat{q}$ denote the prescribed displacement, traction, excess pore pressure, and fluid flux, respectively.
The initial conditions are given by
\begin{align}
    \bm{u}(\bm{x}, 0) = \bm{u}_0(\bm{x}) &\quad\text{in}\;\; \Omega, \\
    p(\bm{x}, 0) = p_0(\bm{x}) &\quad\text{in}\;\; \Omega.
    \label{eq:initial-condition}
\end{align}
The resulting initial--boundary value problem can be solved using standard two-field mixed finite elements.
The discretization details are omitted for brevity.

\section{DeepONet surrogate modeling} 
\label{sec:methodology}

In this section, we develop a DeepONet-based surrogate modeling framework for poroelasticity with random permeability fields. 
The objective is to approximate the solution operator that maps a realization of the permeability field to the corresponding poroelastic response, such as displacement and excess pore pressure fields. 
Let $k(\bm{x}; \omega)$ denote the realization of the input random permeability field, and let $f(\bm{x}, t; \omega)$ denote the corresponding poroelastic response. 
The underlying operator is expressed as
\begin{equation}
    \mathcal{G}: k(\bm{x}; \omega) \mapsto f(\bm{x}, t; \omega),
    \label{eq:poroelasticity-operator}
\end{equation}
where $\mathcal{G}$ represents the mapping induced by the governing initial--boundary value problem described in Section~\ref{sec:poroelasticity}.

To construct an efficient surrogate for $\mathcal{G}$ based on DeepONet, we present a three-stage modeling strategy in the following.
First, we nondimensionalize the governing equations to improve numerical stability and facilitate generalization across varying physical scales. 
Second, we perform dimensionality reduction of the input permeability field using the K--L expansion, which enables extracting dominant modes of variability while compressing the stochastic input space. 
Third, we develop a tailored DeepONet architecture and train it using a two-step training procedure.

\subsection{Nondimensionalization of governing equations} 

Nondimensionalization plays a critical role in constructing stable and generalizable surrogate models, particularly in neural network-based approximations~\cite{amini2022physics, haghighat2022physics, xie2022data}. 
By rescaling variables and parameters into dimensionless form, we reduce numerical stiffness, normalize data magnitudes, and facilitate consistent learning across a range of parameter values.

We consider a two-dimensional poroelastic problem ($d = 2$) and introduce characteristic scaling quantities $T^*$, $s^*$, $k^*$, $u^*$, $p^*$, $t^*$, and $q^*$, corresponding to characteristic time, length, permeability, displacement, pressure, traction, and flux scales, respectively. 
The nondimensionalized variables or parameters are defined as
\begin{equation}
\begin{aligned}
    \bar{t} &\coloneqq \frac{t}{T^*}, \quad
    \bar{x} \coloneqq \frac{x}{s^*}, \quad
    \bar{z} \coloneqq \frac{z}{s^*}, \quad
    \bar{l}_x \coloneqq \frac{l_x}{s^*}, \quad
    \bar{l}_z \coloneqq \frac{l_z}{s^*}, \\
    \bar{k} &\coloneqq \frac{k}{k^*}, \quad
    \bar{\bm{u}} \coloneqq \frac{\bm{u}}{u^*}, \quad
    \bar{p} \coloneqq \frac{p}{p^*}, \quad
    \bar{\bm{\sigma}} \coloneqq \frac{\bm{\sigma}}{t^*}, \quad
    \bar{\bm{q}} \coloneqq \frac{\bm{q}}{q^*},
\end{aligned}
\label{eq:nd-variables}
\end{equation}
where $\bar{\circ}$ denotes the nondimensionalized version of a given variable or parameter $\circ$. 

To determine the scaling parameters, we proceed as follows:
\begin{enumerate}
    \item We set the reference length $s^*$, reference permeability $k^*$, and one loading-related scale---either the reference traction $\hat{t}^*$ or the reference flux $\hat{q}^*$---depending on the loading type in the problem.
    \item Using dimensional analysis of the poroelastic governing equations and constitutive relations, we express the displacement and pressure scales $u^*$ and $p^*$ in terms of the loading scale:
    \begin{equation}
    \begin{cases}
        u^* = A \dfrac{\mu_f {s^*}^2}{k^* E} \hat{q}^*, \quad
        p^* = B \dfrac{\mu_f s^*}{k^*} \hat{q}^* & \text{(for flux loading)}, \vspace{0.75em}\\
        u^* = A \dfrac{s^*}{E} \hat{t}^*, \quad
        p^* = B \hat{t}^* & \text{(for traction loading)},
    \end{cases}
    \label{eq:scaling-up}
    \end{equation}
    where $A$ and $B$ are dimensionless constants dependent on Poisson's ratio $\nu$.
    \item Finally, we define the characteristic time scale as
    \begin{equation}
        T^* = \frac{A}{B} \frac{\mu_f {s^*}^2}{k^* E}.
        \label{eq:scaling-time}
    \end{equation}
\end{enumerate}

In this work, we focus exclusively on spatial variability in the permeability field, while assuming that other elasticity and fluid parameters---namely Young's modulus $E$, Poisson’s ratio $\nu$, and fluid viscosity $\mu_f$---are constants. For notational simplicity, we omit the bars in subsequent sections; all quantities are assumed nondimensional unless stated otherwise.

\subsection{Dimension reduction using Karhunen--Lo\'{e}ve expansion} 

The poroelasticity operator defined in Eq.~\eqref{eq:poroelasticity-operator} maps a permeability field to the corresponding poroelastic response. 
In a discretized form, this operator can be expressed as
\begin{equation}
    \mathcal{G}: \mathbb{R}^{N_s} \ni \bm{k}(\omega) \mapsto f(\bm{x}, t; \omega),
    \label{eq:poroelasticity-operator-discretization}
\end{equation}
where 
\begin{equation}
    \bm{k}(\omega) = \left(k(\bm{x}_1; \omega), \cdots, k(\bm{x}_{N_s}; \omega)\right)
\end{equation}
denotes the discretized permeability field evaluated at $N_s$ spatial points $\{\bm{x}_i\}_{i=1}^{N_s}$. 
While such representation allows for high-resolution modeling of spatial variability, the resulting input dimensionality can be prohibitively large for surrogate modeling.

To reduce the dimensionality of the input, we employ the Karhunen--Lo\`eve (K--L) expansion, which represents a second-order random field in terms of orthogonal spatial basis functions and uncorrelated random variables. Specifically, we express the log-permeability field $\kappa(\bm{x}; \omega) = \ln k(\bm{x}; \omega)$ as
\begin{equation}
    \kappa^m(\bm{x}; \omega) = \mu_\kappa + \sum_{j = 1}^m \sqrt{\lambda_j} \, e_j(\bm{x}) \, \xi_j(\omega),
    \label{eq:kl-log-permeability}
\end{equation}
where $\mu_\kappa$ is the mean, $\lambda_j$ and $e_j$ are the eigenvalues and eigenfunctions of the covariance kernel $C(\bm{x}_1, \bm{x}_2)$, and $\xi_j(\omega) \sim \mathcal{N}(0, 1)$ are uncorrelated standard Gaussian random variables. 
The eigenfunctions satisfy the homogeneous Fredholm integral equation
\begin{equation}
    \int_\Omega C(\bm{x}, \bm{x}') \, e_j(\bm{x}') \, \mathrm{d}\bm{x}' = \lambda_j \, e_j(\bm{x}),
    \label{eq:fredholm}
\end{equation}
and are computed using the integral method~\cite{wang2008karhunen}. 
The eigenvalues are ordered as $\lambda_1 \geq \lambda_2 \geq \cdots \geq \lambda_{N_s}$.

To reconstruct the permeability field, we apply the exponential map:
\begin{equation}
    k^m(\bm{x}_i; \omega) = \exp \left( \mu_\kappa - \kappa^* + \sum_{j = 1}^m \sqrt{\lambda_j} \, e_j(\bm{x}_i) \, \xi_j(\omega) \right),
    \label{eq:k-l-expansion-truncated}
\end{equation}
where $\kappa^* = \ln(k^*)$ is a scaling constant for nondimensionalization. The truncated form $k^m$ defines an approximate permeability realization evaluated at $\{\bm{x}_i\}_{i=1}^{N_s}$:
\begin{equation}
    \bm{k}^m(\omega) = \left(k^m(\bm{x}_1; \omega), \cdots, k^m(\bm{x}_{N_s}; \omega)\right).
\end{equation}

\begin{figure}[htbp]
    \centering
    \includegraphics[width=0.95\textwidth]{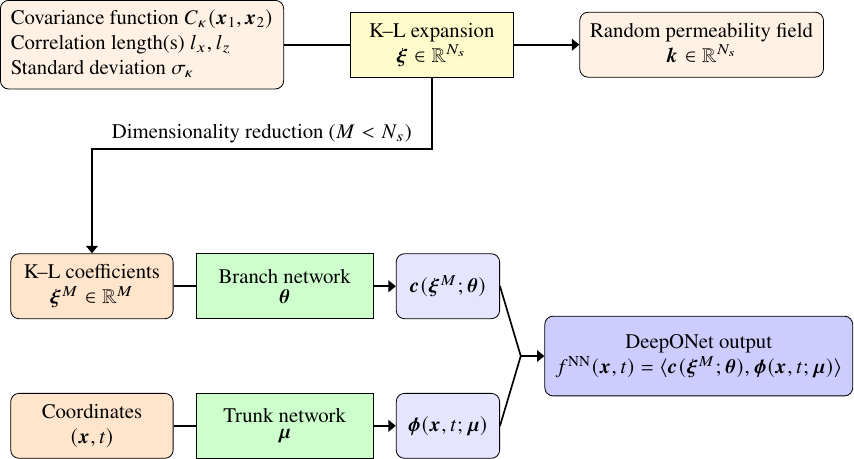}
    \caption{DeepONet surrogate modeling framework for poroelasticity with random permeability fields. A separate DeepONet is trained for each nondimensionalized primary variable and each set of permeability statistics. The predicted output is denoted by $f^{\mathrm{NN}} \in \{u_x^{\mathrm{NN}}, u_z^{\mathrm{NN}}, p^{\mathrm{NN}}\}$.}
    \label{fig:deeponet-surrogate-modeling}
\end{figure}

Accordingly, the operator in Eq.~\eqref{eq:poroelasticity-operator-discretization} becomes
\begin{equation}
    \mathcal{G}: \mathbb{R}^{N_s} \ni \bm{k}^m(\omega) \mapsto f^m(\bm{x}, t; \omega),
    \label{eq:operator-discretization-truncated}
\end{equation}
where $f^m$ is the poroelastic response corresponding to the truncated permeability field. The exact response corresponds to $m = N_s$, that is,
\begin{equation}
    f(\bm{x}, t; \omega) = f^{N_s}(\bm{x}, t; \omega).
\end{equation}
To further reduce the input dimension, we retain only the first $M < N_s$ terms in the expansion, leading to
\begin{equation}
    \bm{\xi}^M(\omega) = \left(\xi_1(\omega), \cdots, \xi_M(\omega)\right) \in \mathbb{R}^M,
    \label{eq:kl-coefficients}
\end{equation}
which serve as a low-dimensional representation of the random input. 
The approximation is acceptable if the associated error
\begin{equation}
    \left\| f(\bm{x}, t; \omega) - f^M(\bm{x}, t; \omega) \right\|^2_{\Omega \times \mathcal{T}} \ll 1
    \label{eq:truncation-error}
\end{equation}
is sufficiently small for the application of interest. Because $\bm{k}^M(\omega)$ is uniquely determined by $\bm{\xi}^M(\omega)$ for fixed mean and covariance, we define the reduced-order operator
\begin{equation}
    \mathcal{G}': \mathbb{R}^M \ni \bm{\xi}^M \mapsto f,
    \label{eq:target-operator-final}
\end{equation}
which maps the K--L coefficients directly to the poroelastic response. 
This formulation enables a compact and structured input representation suitable for neural operator learning using DeepONet.
The DeepONet framework is illustrated in Fig.~\ref{fig:deeponet-surrogate-modeling}. 

\subsection{DeepONet architecture and training} 
In this work, we construct a DeepONet-based surrogate that approximates the operator $\mathcal{G}' $~\eqref{eq:target-operator-final}, mapping the principal K--L expansion coefficients of the random permeability field to the poroelastic response. Specifically, we represent the surrogate operator as
\begin{equation} 
    \mathcal{G}_{\mathrm{NN}}[\bm{\xi}^M](\bm{x}, t) = \langle \bm{c}(\bm{\xi}^M; \bm{\theta}), \bm{\phi}(\bm{x}, t; \bm{\mu}) \rangle,
    \label{eq:deeponet}
\end{equation}
where $\bm{c}: \mathbb{R}^M \rightarrow \mathbb{R}^K$ is the branch network parameterized by $\bm{\theta}$, and $\bm{\phi}: \mathbb{R}^{d + 1} \rightarrow \mathbb{R}^K$ is the trunk network parameterized by $\bm{\mu}$. 
Here, $K$ is the number of basis functions, and $\langle \cdot, \cdot \rangle$ denotes the dot product in $\mathbb{R}^K$.
The branch network encodes the latent representation of the input $\bm{\xi}^M$, producing a coefficient vector $\bm{c} = (c_1, \cdots, c_K)$. 
The trunk network outputs $\bm{\phi} = (\phi_1, \cdots, \phi_K)$ are interpreted as spatiotemporal basis functions, each associated with a scalar coefficient from the branch output. Their dot product provides the prediction of the DeepONet, $f^{\mathrm{NN}}(\bm{x}, t)$.

To handle the different response characteristics of each primary variable, we implement separate DeepONets for each component $f \in \{u_x, u_z, p\}$. This design not only improves flexibility in training but also reduces computational cost per network. Both the branch and trunk networks are constructed using fully connected feed-forward neural networks with hyperbolic tangent activation functions.
The network architecture is specified by layer widths. 
The branch network uses the architecture $\bm{n}_b = (M, \cdots, K)$, and the trunk network uses $\bm{n}_t = (d + 1, \cdots, K - 1)$, where each component denotes the number of neurons in that layer. 
Note that the trunk network has an output with dimension $K - 1$; we manually add a constant basis function after training to ensure completeness.

To train the DeepONet, we adopt the two-step training method introduced by Lee and Shin~\cite{lee2024training}, which has been shown to improve training stability and predictive accuracy relative to standard training. 
Instead of simultaneously optimizing both the trunk and branch networks, the two-step training method separates the training into sequential sub-problems that decouple basis function learning from coefficient learning, reducing optimization complexity and improving generalization.
The training procedure proceeds as follows:
\begin{enumerate}
    \item \textbf{Trunk network training.}  
    We first optimize the trunk network parameters $\bm{\mu}$ and a coefficient matrix $\bm{A}$ by minimizing the loss
    \begin{equation}
        \mathcal{L}_T(\bm{\mu}, \bm{A}) = \left\| \bm{A} \bm{\Phi}(\bm{T}; \bm{\mu})^\mathsf{T} - \bm{F} \right\|_{2, 2}^2,
        \label{eq:trunk-loss}
    \end{equation}
    where $\bm{F} \in \mathbb{R}^{N \times m_\mathrm{y}}$ is the train dataset of output snapshots (\eg~displacement or pressure fields), and $\bm{\Phi}(\bm{T}; \bm{\mu}) \in \mathbb{R}^{m_\mathrm{y} \times K}$ is the trunk network output evaluated at spatiotemporal coordinates $\bm{T}$, discretized with $m_\mathrm{y}$ points. 
    This step treats the trunk network outputs as a learnable functional basis, optimized to minimize global reconstruction error across training samples.
    \item \textbf{QR decomposition and branch target projection.}  
    After trunk network training, we orthogonalize the learned basis using QR decomposition: $\bm{\Phi}^* = \bm{Q}^* \bm{R}^*$, where $\bm{Q}^* \in \mathbb{R}^{m_\mathrm{y} \times K}$ contains orthonormal basis functions and $\bm{R}^* \in \mathbb{R}^{K \times K}$ is upper triangular. 
    We then compute the projected coefficients
    \begin{equation}
        \bm{B}^* = \bm{A}^* \bm{R}^{*\mathsf{T}},
        \label{eq:branch-target}
    \end{equation}
    which define the targets for the branch network. 
    This Gram-Schmidt orthonomalization process ensures that the branch learns coefficients aligned with an orthonormal basis, simplifying the learning task and improving conditioning.
    \item \textbf{Branch network training.}  
    Finally, we train the branch network by minimizing the loss
    \begin{equation}
        \mathcal{L}_B(\bm{\theta}) = \left\| \bm{C}(\bm{\Xi}; \bm{\theta}) - \bm{B}^* \right\|_{2, 2}^2,
        \label{eq:branch-loss}
    \end{equation}
    where $\bm{\Xi} \in \mathbb{R}^{N \times M}$ is the input dataset of K--L expansion coefficients, and $\bm{C}(\bm{\Xi}; \bm{\theta}) \in \mathbb{R}^{N \times K}$ denotes the branch network output. 
    Since the targets are already projected, this step becomes a well-conditioned regression problem.
\end{enumerate}

For optimization, we employ a hybrid strategy~\cite{rathore2024challenges} that combines AdamW~\cite{loshchilov2017decoupled} and L-BFGS~\cite{liu1989limited}. 
We first train each network with AdamW for 5,000 epochs using a mini-batch size of $N / 4$. 
The initial learning rate is set to $10^{-3}$, with exponential decay: $0.99$ per 100 iterations for the trunk network, and $0.98$ for the branch network. 
Afterward, we switch to L-BFGS for fine-tuning with a maximum of 10,000 iterations and a stopping tolerance of $10^{-10}$, as implemented in \texttt{jaxopt}~\cite{jaxopt_implicit_diff}.

\section{Numerical examples} 
\label{sec:numerical-examples}

In this section, we evaluate the proposed surrogate modeling framework using two representative poroelasticity problems: (1) soil consolidation and (2) ground subsidence induced by groundwater extraction. 
The first problem is driven by mechanical (traction) loading, while the second is driven by hydraulic (flux) loading. 
In both cases, the objective is to learn the nonlinear operator that maps random permeability fields to corresponding transient poroelastic responses.

The numerical experiments are designed to assess the performance of surrogate models across a range of permeability statistics representative of subsurface heterogeneity encountered in geotechnical and hydrogeologic systems. 
This parameter space spans a broad spectrum of subsurface conditions, enabling a systematic evaluation of surrogate robustness with respect to spatial variance and correlation structure.

Each example follows the same modeling workflow. 
We first define the physical configuration, nondimensionalize the governing equations, and model the permeability field as a spatially correlated random process. 
We then generate input--output datasets using finite element simulations and train DeepONet to approximate the associated solution operator. 
Using evaluations on a test dataset, we select suitable network architectures and assess generalization across permeability statistics. 
Finally, we compare the computational cost of DeepONet and FEM to determine the crossover point at which surrogate-based inference becomes more efficient.

All FEM simulations are performed using the \texttt{FEniCSx} library~\cite{alnaes2014unified, barrata2023dolfinx} on a single core of an AMD Ryzen Threadripper PRO 5955WX CPU. 
DeepONet models are implemented using \texttt{JAX}~\cite{jax2018github} and \texttt{FLAX}~\cite{flax2020github}, and trained on a single NVIDIA GeForce RTX 4090 GPU.

\subsection{Soil consolidation} 
\label{sec:soil-consolidation}

Our first example is a soil consolidation problem, in which the classical one-dimensional Terzaghi formulation is extended to incorporate spatial random permeability fields. 
This type of problem is of fundamental importance in geotechnical engineering and has been the subject of probabilistic analysis in the literature (\eg~\cite{huang2010probabilistic, cheng2017consolidation}).
The objective is to predict the nondimensionalized vertical displacement $u_z(\bm{x}, t)$ and excess pore pressure $p(\bm{x}, t)$. 
Horizontal displacement is omitted, as the problem geometry, loading conditions, and underlying physics remain effectively one-dimensional, apart from the spatial variability introduced by the permeability field.

\begin{figure}[htbp]
    \centering
    \includegraphics[width=0.5\textwidth]{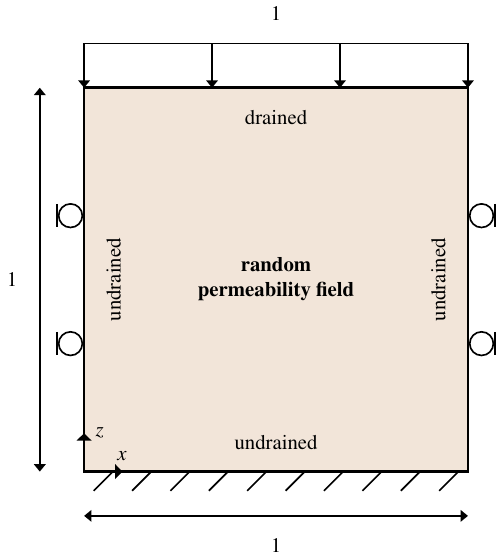}
    \caption{Soil consolidation: Problem setup in nondimensional form.}
    \label{fig:consolidation-setup}
\end{figure}

The problem domain is a unit square representing a saturated soil layer subjected to surface loading. 
At $t = 0$, a uniform vertical stress of magnitude $t_0$ is applied to the top surface. 
Hydraulic boundary conditions impose zero excess pore pressure (drained) on the top edge and no flow (undrained) on the remaining boundaries. Mechanical conditions consist of roller constraints on the lateral edges (zero horizontal displacement and vertical traction) and a fully fixed base.
The Poisson’s ratio is set to $\nu = 0.4$.
To reduce parameter dependence and improve numerical conditioning, we nondimensionalize the governing equations following Terzaghi’s classical consolidation framework. 
With $t^* = t_0$ as the reference stress, the scaling parameters are defined as
\begin{equation}
    T^* = \frac{L^2}{c_v}, \quad s^* = L, \quad k^* = \exp(\mu_\kappa), \quad u^* = m_v L t_0, \quad p^* = t_0,
\end{equation}
where $m_v := \frac{(1 - 2 \nu)(1 + \nu)}{E(1 - \nu)}$ is the coefficient of volume compressibility, and $c_v := \frac{k^*}{\mu_f} \frac{E(1 - \nu)}{(1 - 2 \nu)(1 + \nu)}$ is the coefficient of consolidation. 
The nondimensionalized configuration is illustrated in Fig.~\ref{fig:consolidation-setup}.

To model spatial variability in permeability, we generate random fields using a K--L expansion with a Gaussian covariance function:
\begin{equation}
    C_\kappa(\bm{x}_1, \bm{x}_2) = \sigma_\kappa^2 \exp \left( -\left[ \left( \frac{x_1 - x_2}{l_x} \right)^2 + \left( \frac{z_1 - z_2}{l_z} \right)^2 \right] \right).
\end{equation}
This form captures anisotropic spatial correlation characteristics of sedimentary formations, where vertical layering leads to shorter correlation lengths in the vertical direction~\cite{rehfeldt1992field, phoon1999characterization, wang2022probabilistic}. 
Unless otherwise stated, we adopt the baseline parameter set $(\sigma_\kappa, l_x, l_z) = (1.5, 0.25, 0.125)$.
To construct training data, we draw 10,000 K--L  coefficient samples using Latin hypercube sampling (LHS)~\cite{mckay2000comparison}, allocating 8,000 for training and 2,000 for testing.  
Each sample is evaluated using FEM on a unit square domain $\Omega = \{(x, z) \,|\, 0 \leq x \leq 1,0 \leq z \leq 1\}$ discretized with structured triangular elements ($\Delta x = \Delta z = 0.05$). 
The spatial discretization uses Taylor--Hood elements, where the displacement field is interpolated using quadratic shape functions and the pressure field is interpolated using linear shape functions.
We define the temporal domain as $\mathcal{T} = \{t \,|\, 0 < t \leq 1\}$ and discretized with a uniform time step $\Delta t = 0.01$.

\begin{figure}[htbp]
    \centering
    \subfloat[Vertical displacement $u_z$]
    {\includegraphics[width = 0.45\textwidth]{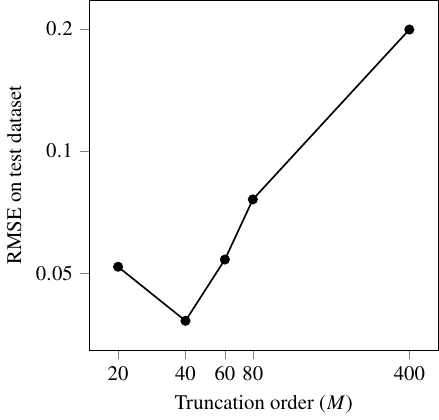}} $\quad$  
    \subfloat[Excess pore pressure $p$]
    {\includegraphics[width = 0.45\textwidth]{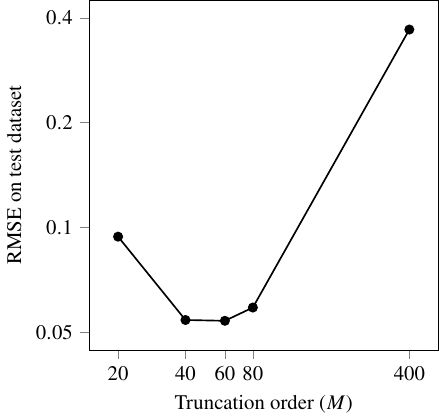}}
    \caption{Soil consolidation: Test RMSE for truncation orders $M \in \{20, 40, 60, 80, 400\}$ with $(\sigma_\kappa, l_x, l_z) = (1.5, 0.25, 0.125)$. Left: vertical displacement $u_z$. Right: excess pore pressure $p$.}
    \label{fig:consolidation-error}
\end{figure}

Based on the FEM-generated dataset, we construct a DeepONet that maps low-dimensional K--L inputs to transient poroelastic responses. 
The branch network receives the truncated K--L coefficient vector $\bm{\xi}^M$, and the trunk network is evaluated on a uniform grid.
The number of basis functions $K$ is chosen to be 1--2 times the numerical rank of the output matrix $\bm{U}$, determined using a relative singular value threshold of $0.01 \times \max(N, m_y)$; singular values below this threshold are considered negligible for basis selection.
The output consists of the evaluated primary variables at spatial nodes $\{(x_i, z_i)\ |\ x_i \in \{0.0,0.1,\dots, 0.9, 1.0\},z_i \in \{0.0, 0.1, \dots, 0.9, 1.0\}\}$ and temporal nodes $t_i \in \{0.1, 0.2, \dots, 1.0\}$.
The trunk network is trained independently for each output variable. 
The trunk network architectures are configured as $\bm{n}_t = (3, 256, 256, 256)$ for vertical displacement $u_z$ and $\bm{n}_t = (3, 64, 64, 32)$ for excess pore pressure $p$. 
The branch network adopts the architecture $\bm{n}_b = (M, 64, 64, 64, 64, K)$, with truncation orders $M \in \{20, 40, 60, 80, 100, 400\}$ considered. 
Predictive accuracy is quantified using the relative test error
\begin{equation}
    \text{RMSE} = \frac{\| \bm{F}^{\text{NN}} - \bm{F} \|^2_{2,2}}{\| \bm{F} \|^2_{2,2}},
\end{equation}
where $\bm{F}$ and $\bm{F}^{\text{NN}}$ are the FEM and DeepONet outputs, respectively. 
As shown in Fig.~\ref{fig:consolidation-error}, the optimal truncation orders are $M = 40$ for $u_z$ and $M = 60$ for $p$, and we adopt these values for the remainder of the problem.

To evaluate the robustness of the surrogate across different variability levels, we test the trained DeepONet surrogate models using the following permeability statistics: $(\sigma_\kappa, l_x, l_z) \in \{1.5, 1.0, 0.5\} \times \{0.25, 0.5\} \times \{0.125, 0.25\}$. 
Table~\ref{tab:consolidation-statistical-parameter-prediction-accuracy} reports the resulting test RMSEs for both primary variables. 
As expected, prediction errors increase with stronger spatial variability and shorter correlation lengths. 
Excess pore pressure is consistently more difficult to approximate, reflecting its heightened sensitivity to fine-scale permeability features.
\begin{table}[htbp]
    \renewcommand{\arraystretch}{1.5}
    \centering
    \resizebox{\textwidth}{!}{
    \begin{tabular}{lllllll}
        \toprule
        & \multicolumn{3}{l}{\textbf{Vertical displacement} ($u_z$)} & \multicolumn{3}{l}{\textbf{Excess pore pressure} ($p$)} \\
        \cline{2-7}
        &$\sigma_\kappa = 1.5$&$\sigma_\kappa = 1.0$&$\sigma_\kappa = 0.5$& 
       $\sigma_\kappa = 1.5$&$\sigma_\kappa = 1.0$&$\sigma_\kappa = 0.5$\\
        \midrule
       $(l_x, l_z) = (0.25, 0.125)$& 
       $3.83 \times 10^{-2}$&$1.13 \times 10^{-2}$&$2.56 \times 10^{-3}$&
       $5.36 \times 10^{-2}$&$2.34 \times 10^{-2}$&$7.42 \times 10^{-3}$\\
       $(l_x, l_z) = (0.25, 0.25)$& 
       $2.03 \times 10^{-2}$&$6.69 \times 10^{-3}$&$1.84 \times 10^{-3}$& 
       $4.72 \times 10^{-2}$&$1.94 \times 10^{-2}$&$5.76 \times 10^{-3}$\\
       $(l_x, l_z) = (0.5, 0.125)$& 
       $2.03 \times 10^{-2}$&$6.20 \times 10^{-3}$&$1.75 \times 10^{-3}$& 
       $3.28 \times 10^{-2}$&$1.57 \times 10^{-2}$&$5.17 \times 10^{-3}$\\
       $(l_x, l_z) = (0.5, 0.25)$& 
       $9.41 \times 10^{-3}$&$3.49 \times 10^{-3}$&$1.38 \times 10^{-3}$& 
       $2.64 \times 10^{-2}$&$1.36 \times 10^{-2}$&$4.10 \times 10^{-3}$\\
        \bottomrule
    \end{tabular}} 
    \caption{Soil consolidation: Test RMSE for various permeability statistics $(\sigma_\kappa, l_x, l_z)$.}
    \label{tab:consolidation-statistical-parameter-prediction-accuracy}
\end{table}

\begin{figure}[htbp]
    \centering
    \includegraphics[width=0.9\textwidth]{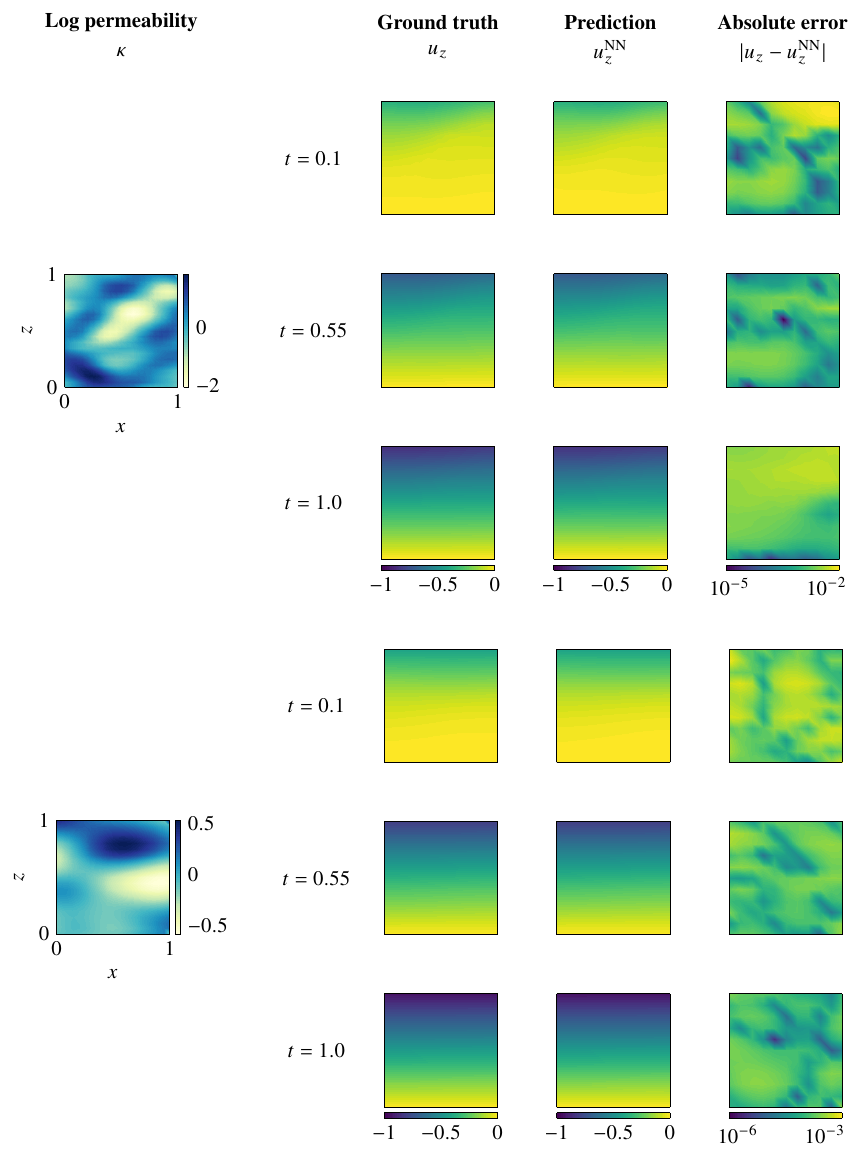}
    \caption{Soil consolidation: Comparison between FEM and DeepONet predictions for vertical displacement. 
    Top: high variance and short correlation lengths $(\sigma_\kappa, l_x, l_z) =  (1.5, 0.25, 0.125)$. 
    Bottom: low variance and long correlation lengths $(\sigma_\kappa, l_x, l_z) = (0.5, 0.5, 0.25)$.}
    \label{fig:consolidation-displacement-z}
\end{figure}

\begin{figure}[htbp]
    \centering
    \includegraphics[width=0.9\textwidth]{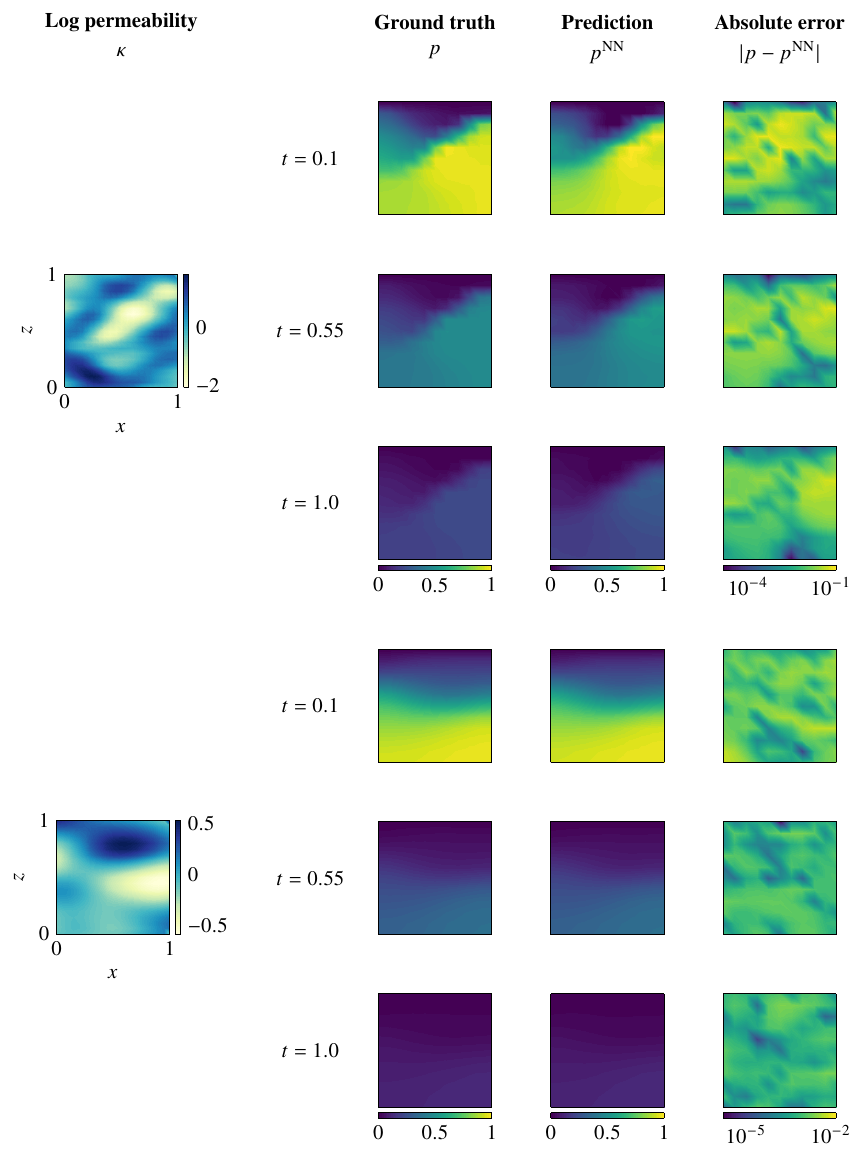}
    \caption{Soil consolidation: Comparison between FEM and DeepONet predictions for excess pore pressure. 
    Top: high variance and short correlation lengths $(\sigma_\kappa, l_x, l_z) = (1.5, 0.25, 0.125)$. 
    Bottom: low variance and long correlation lengths $(\sigma_\kappa, l_x, l_z) =  (0.5, 0.5, 0.25)$.}
    \label{fig:consolidation-pressure}
\end{figure}

To assess predictive accuracy in the domain, we compare DeepONet outputs to FEM solutions at $t = 0.1$, $0.55$, and $1.0$, where $t=0.55$ is an interpolation point not seen during training. 
Figures~\ref{fig:consolidation-displacement-z} and~\ref{fig:consolidation-pressure} show predicted fields and pointwise absolute errors for two representative settings: high variance with short correlation lengths $(\sigma_\kappa, l_x, l_z) = (1.5, 0.25, 0.125)$ and low variance with long correlation lengths $(\sigma_\kappa, l_x, l_z) = (0.5, 0.5, 0.25)$.
The results show that the surrogate accurately simulates both spatial and temporal variations across contrasting permeability regimes.

We compare the computational efficiency of the DeepONet surrogate against the finite element baseline under the same hardware configuration used for data generation and training. 
The total FEM runtime for generating 8,000 training samples is $1.27 \times 10^4$ seconds. 
DeepONet training requires $7.03 \times 10^2$ seconds for vertical displacement $u_z$ and $1.99 \times 10^3$ seconds for excess pore pressure $p$.
Once trained, inference is highly efficient, requiring less than $0.05$ seconds per sample.
To quantify the point at which surrogate modeling becomes advantageous, we define the crossover simulation count $N_c$ as
\begin{equation}
    N_c = N_{\text{tr}} \left(1 + \frac{T_T}{T_F} \right),
    \label{eq:crossover-simulation-count}
\end{equation}
where $N_{\text{tr}}$ denotes the number of training samples, $T_F$ is the total FEM runtime, and $T_T$ is the total training time for the DeepONet. 
The cost of inference and post-processing including QR decomposition is omitted due to their negligible contribution. 
Based on measured runtimes, the crossover thresholds are $N_c = 8,442$ for $u_z$ and $N_c = 9,246$ for $p$. 
These thresholds can be further reduced by decreasing the training size or optimizing network training \cite{ainsworth2022active}.

\subsection{Ground subsidence induced by groundwater extraction} 
\label{sec:subsidence}

We next consider ground subsidence induced by groundwater extraction. 
The objective is to train DeepONet to predict the horizontal displacement $u_x(\bm{x}, t)$, vertical displacement $u_z(\bm{x}, t)$, and excess pore pressure $p(\bm{x}, t)$.

\begin{figure}[htbp]
    \centering
    \includegraphics[width = \linewidth]{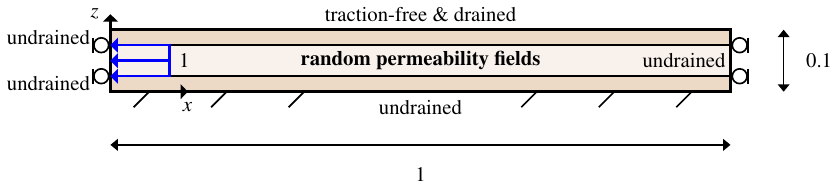}
    \caption{Ground subsidence: Problem setup in nondimensional form.}
    \label{fig:subsidence-setup}
\end{figure}

The spatial domain is defined as a rectangle of width $W$ and height $H$, with $H = 0.1 W$, representing a laterally extensive confined aquifer. 
The domain consists of three horizontal layers: the top and bottom layers, each of thickness $0.3H$, are assigned constant low permeability, while the middle layer, of thickness $0.4H$, contains heterogeneous high-permeability fields. 
The mean log-permeability in the middle layer is set to be 3 units higher than that in the outer layers, consistent with typical contrasts between clay and sand formations~\cite{alghamdi2020bayesian, haagenson2020generalized}.
Zero initial stress and pressure conditions are assumed. 
The mechanical boundary conditions are specified as follows: zero traction on the top boundary, zero horizontal displacement and vertical traction on the lateral boundaries, and zero displacement on the bottom boundary. 
For hydraulic conditions, a constant specific discharge of magnitude $q_0$ is prescribed on the left boundary of the middle layer. 
In addition, zero excess pore pressure is imposed on the top boundary, and zero discharge is applied on the remaining boundaries. 
The Poisson’s ratio is set as $\nu = 0.25$.
The problem is nondimensionalized in a way similar to the confined aquifer solution under plane stress presented in Section 5.5.6 of Verruijt~\cite{verruijt2016}. 
Using $q^* = q_0$ as the characteristic discharge, the scaling parameters are defined as
\begin{equation} 
    T^* = \left(\frac{W}{H} \right) \frac{W^2}{c_v'}, \quad s^* = W, \quad k^* = k_d, \quad u^* = m_v' \frac{\mu_f WH}{k^*} q^*, \quad p^* = \frac{\mu_f W}{k^*} q^* 
    \label{eq:subsidence-scaling-parameter}
\end{equation}
where $k_d$ denotes the permeability at the fluid extraction boundary, $m_v' = \frac{2 (1 - 2 \nu)(1 + \nu)}{E}$, and $c_v' = \frac{k^*}{\mu_f} \frac{E}{2(1 - 2 \nu)(1 + \nu)}$. The nondimensionalized problem setup is illustrated in Fig.~\ref{fig:subsidence-setup}.

We model random permeability fields using the K--L expansion with a Gaussian covariance function
\begin{equation}
    C_\kappa(\bm{x}_1, \bm{x}_2) = \sigma_\kappa^2 \exp \left[ -\left( \frac{x_1 - x_2}{l_x} \right)^2 \right],
\end{equation}
where spatial variability is introduced only in the horizontal direction, in a way similar to the setup in Alghamdi \etal~\cite{alghamdi2020bayesian}. 
In the following, we consider $(\sigma_\kappa, l_x) = (1.5, 0.125)$ unless stated otherwise.
We generate 10,000 samples of K--L expansion coefficients using LHS, allocating 8,000 samples for training and 2,000 for testing. 
For each sample, the corresponding realization of the log-permeability field is constructed via the K--L expansion using the specified statistical parameters $(\sigma_\kappa, l_x)$.
The poroelastic forward problem is solved over the spatial domain $\Omega = \{(x, z) \,|\, 0 \leq x \leq 1,0 \leq z \leq 0.1\}$, discretized with $\Delta x = 0.025$ and $\Delta z = 0.01$ on a structured triangular mesh. 
The temporal domain is defined as $\mathcal{T} = \{t \,|\, 0 < t \leq 1\}$ and discretized with a uniform time step $\Delta t = 0.01$.

The DeepONet surrogate model is constructed using the principal K--L expansion coefficients as input features for the branch network, following the setup in Section~\ref{sec:soil-consolidation}. 
The output consists of the evaluated primary variables at spatial nodes $\{(x_i, z_i)\ |\ x_i \in \{0.0, 0.05, \dots, 0.95, 1.0\},z_i \in \{0.0, 0.02, \dots, 0.10\}\}$ and temporal nodes $t_i \in \{0.1, 0.2, \dots, 1.0\}$.
For displacements, the architecture is set as $\bm{n}_t = (3,64,64,64,64)$, and for excess pore pressure, $\bm{n}_t = (3,64,64,64,128)$. The branch network is defined as $\bm{n}_b = (M,64,64,64,64,K)$, where $M$ is the number of retained K--L modes and $K$ is the dimension of the shared latent representation. 
We examine $M \in \{15,20,25,40,80\}$ and select the value that minimizes the RMSE on the test dataset.

\begin{figure}[htbp]
    \centering
    \subfloat[Horizontal displacement $u_x$]
    {\includegraphics[width = 0.3\textwidth]{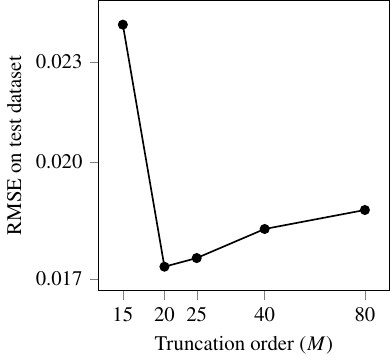}} $\quad$    
    \subfloat[Vertical displacement $u_z$]
    {\includegraphics[width = 0.3\textwidth]{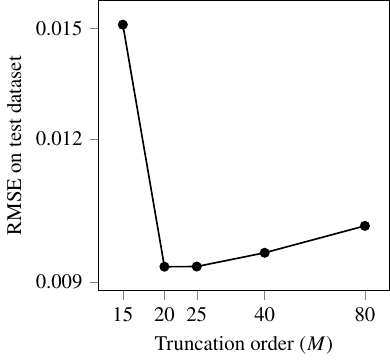}} $\quad$
    \subfloat[Excess pore pressure $p$]
    {\includegraphics[width = 0.3\textwidth]{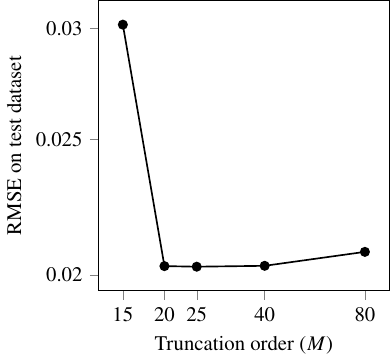}} 
    \caption{Ground subsidence: Test RMSE for truncation orders $M \in \{15,20,25,40,80\}$ for $(\sigma_\kappa,l_x) = (1.5,0.125)$. 
    Left: horizontal displacement $u_x$. 
    Middle: vertical displacement $u_z$. 
    Right: excess pore pressure $p$.}
    \label{fig:subsidence-error}
\end{figure}

Figure~\ref{fig:subsidence-error} shows RMSE on the test dataset for each primary variable and truncation order $M$. 
The results indicate that $M = 20$ yields the lowest test error for all variables. 
This corresponds to one-quarter of the number of horizontal discretization points used for the log-permeability field $\kappa$. 
We adopt this truncation level for the remainder of the problem.

To evaluate the robustness of the surrogate across varying variability levels, we test the trained DeepONet surrogate models using the following permeability statistics: $(\sigma_\kappa, l_x) \in \{1.5, 1.0, 0.5\} \times \{0.125, 0.25, 0.5\}$. 
Table~\ref{tab:subsidence-statistical-parameter-prediction-accuracy} reports the resulting test RMSEs for every primary variable. 
Prediction errors tend to increase with stronger spatial variability and shorter correlation lengths. 

\begin{table}[htbp]
    \renewcommand{\arraystretch}{1.5}
    \centering
    \resizebox{\textwidth}{!}{
        \begin{tabular}{llllllllll}
        \toprule
        & \multicolumn{3}{l}{\textbf{Horizontal displacement} ($u_x$)} & \multicolumn{3}{l}{\textbf{Vertical displacement} ($u_z$)} & \multicolumn{3}{l}{\textbf{Excess pore pressure} ($p$)} \\
        \cline{2-10}
        & $\sigma_\kappa = 1.5$ & $\sigma_\kappa = 1.0$ & $\sigma_\kappa = 0.5$ 
        & $\sigma_\kappa = 1.5$ & $\sigma_\kappa = 1.0$ & $\sigma_\kappa = 0.5$ 
        & $\sigma_\kappa = 1.5$ & $\sigma_\kappa = 1.0$ & $\sigma_\kappa = 0.5$ \\
        \midrule
        $l_x = 0.125$ 
        & $1.85 \times 10^{-2}$ & $1.33 \times 10^{-2}$ & $8.86 \times 10^{-3}$ 
        & $9.78 \times 10^{-3}$ & $7.22 \times 10^{-3}$ & $4.69 \times 10^{-3}$ 
        & $2.07 \times 10^{-2}$ & $1.58 \times 10^{-2}$ & $9.84 \times 10^{-3}$ \\
        $l_x = 0.25$ 
        & $1.35 \times 10^{-2}$ & $1.08 \times 10^{-2}$ & $5.87 \times 10^{-2}$ 
        & $6.99 \times 10^{-3}$ & $5.05 \times 10^{-3}$ & $3.30 \times 10^{-3}$ 
        & $1.42 \times 10^{-2}$ & $1.07 \times 10^{-2}$ & $6.08 \times 10^{-3}$ \\
        $l_x = 0.5$ 
        & $8.26 \times 10^{-3}$ & $6.52 \times 10^{-3}$ & $4.60 \times 10^{-3}$ 
        & $4.81 \times 10^{-3}$ & $3.90 \times 10^{-3}$ & $2.50 \times 10^{-3}$ 
        & $1.07 \times 10^{-2}$ & $8.00 \times 10^{-3}$ & $4.88 \times 10^{-3}$ \\
        \bottomrule
    \end{tabular}}
    \caption{Ground subsidence: Test RMSE for various permeability statistics $(\sigma_\kappa, l_x)$.}
    \label{tab:subsidence-statistical-parameter-prediction-accuracy}
\end{table}

\begin{figure}[htbp]
    \centering
    \includegraphics[width=0.9\textwidth]{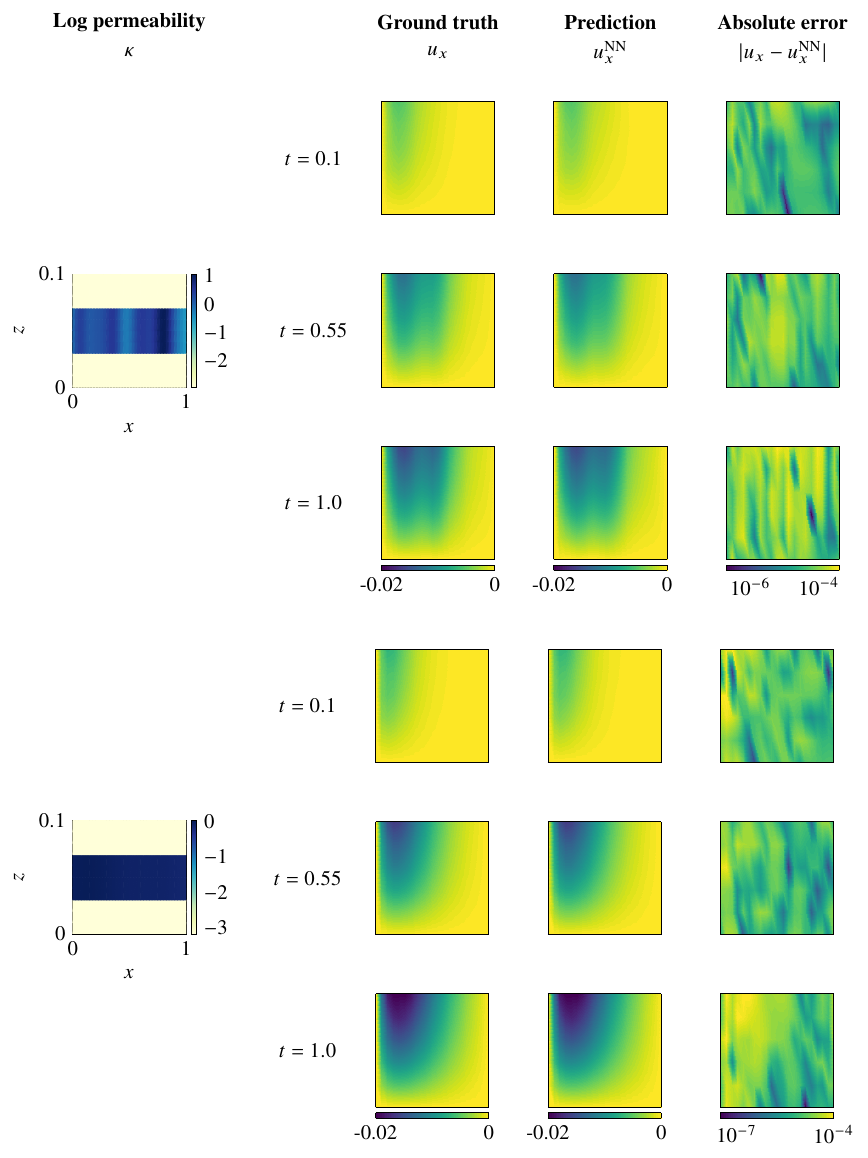}
    \caption{Ground subsidence: Comparison between FEM and DeepONet predictions for horizontal displacement $u_x$. Top: high variance and short correlation length $(\sigma_\kappa,l_x) = (1.5,0.125)$. Bottom: low variance and long correlation length $(\sigma_\kappa,l_x) = (0.5,0.5)$.}
    \label{fig:subsidence-displacement-x}
\end{figure}

\begin{figure}[htbp]
    \centering
    \includegraphics[width=0.9\textwidth]{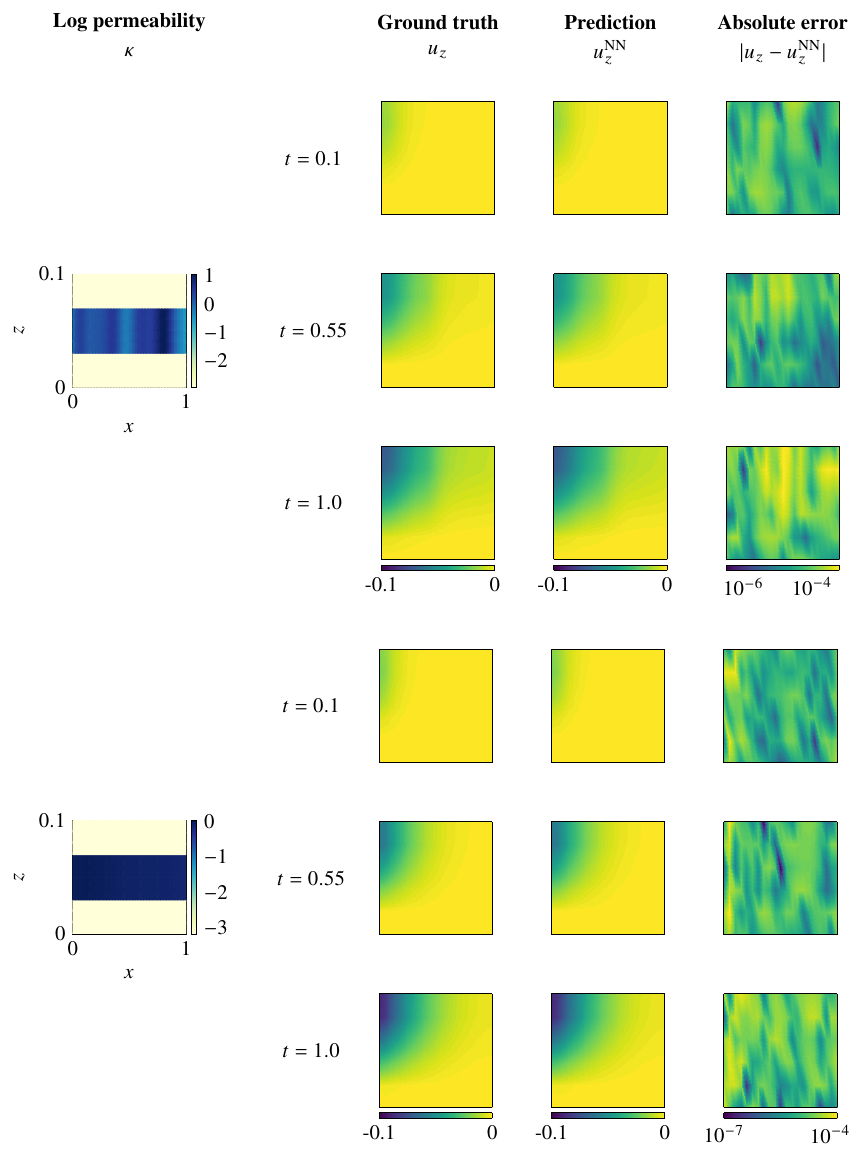}
    \caption{Ground subsidence: Comparison between FEM and DeepONet predictions for vertical displacement $u_z$. Top: high variance and short correlation length $(\sigma_\kappa,l_x) = (1.5,0.125)$. Bottom: low variance and long correlation length $(\sigma_\kappa,l_x) = (0.5,0.5)$.}
    \label{fig:subsidence-displacement-z}
\end{figure}

\begin{figure}[htbp]
    \centering
    \includegraphics[width=0.9\textwidth]{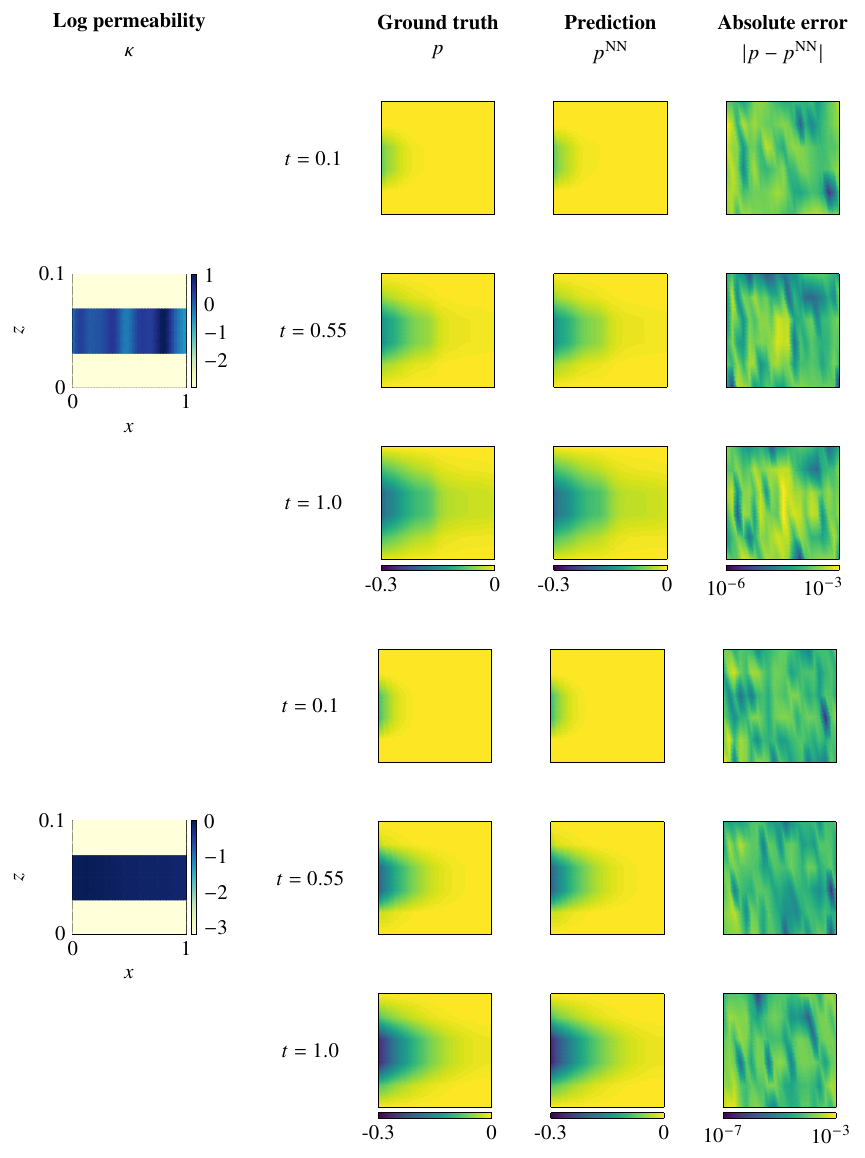}
    \caption{Ground subsidence: Comparison between FEM and DeepONet predictions for excess pore pressure $p$. Top: high variance and short correlation length $(\sigma_\kappa,l_x) = (1.5,0.125)$. Bottom: low variance and long correlation length $(\sigma_\kappa,l_x) = (0.5,0.5)$.}
    \label{fig:subsidence-pressure}
\end{figure}

To assess predictive accuracy in the domain, we compare DeepONet predictions to full-order FEM solutions at $t = 0.1$, $0.55$, and $1.0$, where $t = 0.55$ is an interpolation point not seen during training. 
Figures~\ref{fig:subsidence-displacement-x}, \ref{fig:subsidence-displacement-z}, and~\ref{fig:subsidence-pressure} show the predicted fields and pointwise absolute errors for horizontal displacement $u_x$, vertical displacement $u_z$, and excess pore pressure $p$, respectively. 
Two representative settings are considered: high variance with short correlation length $(\sigma_\kappa,l_x) = (1.5,0.125)$ and low variance with long correlation length $(\sigma_\kappa,l_x) = (0.5,0.5)$. 
Vertical exaggeration is applied to the displacement and excess pore pressure plots for visual clarity. 
The results confirm that the surrogate accurately captures both spatial and temporal variations across contrasting permeability regimes.


We compare computational performance under the same hardware configuration used for data generation and model training. 
The FEM solver requires $6.58 \times 10^3$ seconds to generate 8,000 training samples. 
DeepONet training times are $1.11 \times 10^3$ seconds for $u_x$ and $u_z$, and $1.18 \times 10^3$ seconds for $p$. 
Inference time is less than $0.05$ seconds per sample. 
Based on these measurements, the crossover thresholds, defined in Eq.~\eqref{eq:crossover-simulation-count}, are $N_c = 9,351$ for $u_x$ and $u_z$, and $N_c = 9,432$ for $p$.

\section{Closure} 
\label{sec:closure}

In this paper, we have presented a DeepONet-based surrogate modeling framework for poroelasticity with spatially variable permeability.
The model learns the solution operator mapping realizations of the permeability field to transient displacement and pressure responses. 
To enhance generalization and stability, the framework incorporates nondimensionalization, input compression via truncated K--L expansion, and a two-stage training procedure that decouples the branch and trunk networks.
The surrogate was evaluated on two benchmark problems: soil consolidation under mechanical loading and ground subsidence due to hydraulic extraction. 
In both cases, the model achieved high predictive accuracy across a range of permeability statistics and provided significant computational speedups relative to full-order finite element simulations. 
These results highlight the potential of operator learning to accelerate forward modeling in poroelasticity involving spatial random permeability fields.

Despite these favorable results, several limitations remain. 
The use of K--L truncation, while effective for fields with rapid spectral decay, may be less suitable for input permeability fields characterized by slow-decaying spectra, such as those defined by exponential covariance kernels. 
Also, in its current form, the framework assumes fixed mechanical parameters and a single statistical configuration per network, necessitating retraining for each new setting.
Generalization across broader parametric spaces may be achieved through multi-input operator networks~\cite{jin2022mionet}, which enable the inclusion of both mechanical and statistical parameters as inputs. 
In addition, incorporating architecture-level constraints or physics-informed training objectives may further improve robustness and promote physical consistency. 
These directions will be pursued in future work.

\section*{Acknowledgments}
This work was supported by the National Research Foundation of Korea (NRF) grant funded by the Korean government (MSIT) (No. RS-2023-00209799).
The authors wish to thank Prof. Andy Y.F. Leung and Prof. Taeyong Kim for valuable discussions and insights.

\section*{Data Availability Statement} 
\label{sec:data-availability} 

The data that support the findings of this study are available from the corresponding author upon reasonable request.

\bibliography{references}

\end{document}